# Real-Word Error Correction with Trigrams: Correcting Multiple Errors in a Sentence

**Seyed MohammadSadegh Dashti**

Department of Computer Engineering, Kerman Branch, Islamic Azad University, Kerman, Iran

dashti@iauk.ac.ir

**Abstract**

Spelling correction is a fundamental task in Text Mining. In this study, we assess the real-word error correction model proposed by Mays, Damerau and Mercer and describe several drawbacks of the model. We propose a new variation which focuses on detecting and correcting multiple real-word errors in a sentence, by manipulating a Probabilistic Context-Free Grammar (PCFG) to discriminate between items in the search space. We test our approach on the Wall Street Journal corpus and show that it outperforms Hirst and Budanitsky's WordNet-based method and Wilcox-O'Hearn, Hirst, and Budanitsky's fixed windows size method.

## 1   Introduction

Developing new techniques for auto-correcting errors, especially real-word errors, has proven to be a challenge for researchers. Because of this relatively underdeveloped condition, many existing spelling correction techniques suffer from low accuracy. There are two main tasks in spelling correction: error detection and error correction. Errors found in regular texts are roughly classified into two sets: non-word errors and real-word errors. Non-word errors may appear when a user (e.g. a typist) misspells a word. Because these types of errors do not have a correct spelling, they cannot be verified against a list of dictionary words. As a result, spelling correction programs, such as word processing software solutions, may easily detect such errors and proceed with the error correction task. Nevertheless, real-word errors may occur when a user mistakenly types a correctly spelled word, while intending another word. Many spelling correction programs cannot detect these types of errors because they process words in isolation. Thus, such programs are only capable of detecting non-word errors that are misspelled and are not found in the dictionary database. In contrast, real-word errors may appear when a word with the same pronunciation is typed mistakenly instead of another intended word. These errors may also appear when a word is anomalous in comparison to the other words in the sentence. Real-word errors may also occur when a user tries to correct a non-word error, by using the suggestion list feature in a text processing application. In such cases, a real-word error is generated as the user mistakenly selects the wrong choice among the suggestions to replace the error (Wilcox-O'Hearn and Hirst, 2008). In other cases, spelling correction programs may mistakenly generate a real-word error when trying to correct non-word errors, when the "auto-correct" feature in these programs is enabled (Hirst and Budanitsky, 2005). Pedler (2007) and Hirst and Budanitsky (2005) have conducted comprehensive surveys of real-word spelling correction. Kukich (1992) has also provided an extensive review of the topic. In this study, we propose a new model for auto-correcting multiple real-word errors in a single sentence. This new method uses windows with fixed lengths, which were first introduced by Wilcox-O'Hearn and Hirst (2008). The study explains why this new method detects and auto-corrects real-word errors more effectively, especially when a typist is unskilled. To accomplish these purposes, the study draws on several works in the literature (Wilcox-O'Hearn and Hirst, 2008; Hirst and Budanitsky,

2005), and uses the corpus of Wall Street Journal as a baseline for evaluation. The results, then, can be compared with those of earlier methods (Wilcox-O'Hearn and Hirst, 2008; Hirst and Budanitsky, 2005). This paper is organized based on the following order: Section 2 presents a brief overview of the method (Mays et al., 1991), based on which the present model is developed. Previous variations and improvements on the method (Mays et al., 1991) are explained in section 3. The proposed method is described in section 4. Section 5 presents evaluation and experimental results. An overview of relevant studies is provided in section 6, and finally the conclusion appears in section 7.

## 2 Mays, Damerau and Mercer model
### 2.1 Overview of the model

In this section, the model developed by Mays et al. (1991) is overviewed and some of its important drawbacks are discussed. Mays and colleagues' (1991) model was an instance of the noisy-channel problem. It tried to correct the observed sentence $S$, which had passed through a noisy channel (e.g. a typist) which could generate some errors mistakenly, in the correct sentence $S'$. Mays et al. (1991) considered α a parameter which demonstrated the probability that a word was typed correctly. The remaining fraction (1- α) represented the probability that the word was mistyped as a real-word error. This probability was distributed equally among all the related candidate proposals. Thus, the probability that the target word $w$ typed as $y$ was demonstrated by:

(1) $P(w|y) \begin{cases} \alpha & \text{if } y = w \\ (1-\alpha)/|S_c(w)| & \text{if } y \in S_c(w) \end{cases}$

In equation 1, $S_c(w)$ is the set of spelling variations of $w$ generated by the *ispell* (Kuenning et al., 2004) software. The equation 2, however, was later used to estimate the probability of the all related trigrams and to replace the most probable of them with the original trigram:

(2) $P(S') = \prod_{i=1}^{n} P(w_i \mid w_{i-1} w_{i-2})$

where $n$ does not properly demonstrate end of the sentence, and $n+2$ defines the *EoS* (end of sentence) correctly. Thus, $S'$ is one of the existing sentences in the search space $C(S')$ which maximizes the probability of $P(S'|S) = P(S'). P(S|S')$.

### 2.2 Disadvantages of Mays, Damerau and Mercer's method

One of the disadvantages of this model is the very large size of its trigram model, which is a prerequisite to reach high performance. Another disadvantage is that the model attempts to correct grammatical errors that may be detected and corrected using grammar checkers. The model, however, suffers from an undesirable shortcoming: According to Wilcox-O'Hearn and Hirst (2008), because every member of the search space $C(S)$ is a sentence with one modified word, the method could correct only one real-word error in every sentence. This calculation would be challenging in the case of low α values, in which the source of the noisy channel (e.g. a typist) is likely to make many typos. For instance, in the mistakenly typed piece *"to of thew"*, *t*he user intended to produce *"two of the"*; in this target sentence, $w_1$= *to*, $w_2$= *of*, and $w_3$= *thew*. In this case, Mays and colleagues' (1991) model would generate a search space of related sentences, each containing only one spelling variation of the original sentence words. To be more concise, the search space would include sentences:

$C(S') = \{$
$(S_c(w_1)\ w_2\ w_3). (w_1\ S_c(w_2)\ w_3). (w1\ w2\ Sc(w3))\}$

Furthermore, the model would attempt to estimate the probability of every sentence in the current search space, and to replace the original sentence with one which shows the highest probability. However, because there is no sentence with more than one spelling variation in the search space, no proper replacement for the original sentence could be found.

## 3 Previous variations and improvements on the model

Wilcox-O'Hearn and Hirst (2008) presented a new evaluation of the Mays and colleagues' (1991) model. They tested the model on the corpus of Wall Street Journal. Although the model performed very well in comparison to that of Hirst and Budanitsky (2008), there was still room for improvement. The following sub-sections address the two major attempts that Wilcox-O'Hearn and Hirst, (2008) made to improve the model developed by Mays et al.

## 3.1 Multiple corrections per sentence

The model of Mays et al. (1991) normally made only one correction in every sentence. Wilcox-O'Hearn and Hirst (2008) viewed this limitation as an NP hard problem to include sentences with more than one correction in the search space. They pointed out that such a capability would be useful only in cases where the typist was quite careless, or in other words, when the α value was low. In order to solve this challenge, Wilcox-O'Hearn and Hirst (2008) proposed a solution: Instead of a single sentence from the search space $C(S')$, they chose all the sentences which showed a higher probability value than the original sentence $S$ and used a combination of them.

## 3.2 Using windows of fixed length

The model of Mays et al. (1991) normally used sentences as variable-length units to optimize itself. In contrast, Wilcox-O'Hearn and Hirst (2008) introduced a variation of the model which optimized itself through windows with fixed lengths. In this variation the original boundaries of the sentence were represented by BoS (Beigining-of-Sentence) and EoS (End-of-Sentence). A window with fixed length $d+4$, where $d$ was the span of words, was then used to accommodate all possible trigrams which overlapped with the words in the current span. As a result, the smallest window size would be 5, including three trigrams to estimate the probability of all the spelling variations of the middle word in the span. The method would then move $d$ words to the right and check all the other words in the sentence. As a result, the model made it possible to process multiple corrections in a sentence. If $l$ was the length of the sentence (including sentence markers, *BoS* and *EoS*), then $l−d+1$ iterations were needed to check the sentence completely.

## 4 The improved variation of Mays, Damerau and Mercer's model

Nowadays, contrary to 1980s and early 1990s, computers and electronic devices are widely available to people worldwide. Individuals can use a variety of typing electronic devices. Different people have different levels of attitude, accuracy and speed in typing documents. To reach a good level of practicality, programs should be capable of adjusting themselves to different types of users, ranging from professional typists to unskilled people who use handheld devices. As Wilcox-O'Hearn and Hirst (2008) observe, conducting multiple corrections in a sentence is not always necessary. Yet, their opinion does not seem to be totally true, if one views the problem in the light of real world practice. Consider $α = .9$, which is a very common value for this parameter. This value suggests that out of every ten words typed by the user, one is incorrect on average. In reality, it is likely that a user types two words incorrectly, one after another, or with a distance of two or three words. Since Mays and colleagues' (1991) model is only capable of correcting one error in a sentence, it could not normally help correct multiple real-word errors. As it is already explained, Wilcox-O'Hearn and Hirst (2008) introduced a variation which used fixed windows length. Using windows with fixed-lengths permits multiple corrections per sentence. However, the challenge is that multiple corrections cannot be made in a specific window. As an instance, consider the following sentence taken from the corpus of Wall Street Journal:

*…the Senate to support aim → him [aid] for them [the] Contras…*

This is a false-positive correction example of fixed-window size, where *"aim"* and *"them"* are real word errors, but Wilcox-O'Hearn and Hirst's (2008) fixed-window failed to correct the first error. Yet, even more importantly, it did not detect the second real-word error because the method could only detect and correct one error in per window.

## 4.1 Multiple corrections per window

Our main contribution is to formulate a method which can process multiple corrections in cases where there is more than one error in a window. There are two primary steps to achieve this goal: first generate the members of the search space, and second manipulate the PCFG and discriminate between the existing members of the search space efficiently to find the most probable correction candidate. The next three sub-sections explain the process in detail.

### 4.1.1 Generating the current window's search space

In the model suggested by Mays et al. (1991), the search space includes a set of sentences in which there is no more than one variation. However, in Wilcox-O'Hearn and Hirst's (2008) model, instead of sentences, the algorithm runs on smaller units which are fixed-sized windows. This allows the model to correct the maximum of one error per window in the sentence. However, in practice the model is not still capable of correcting and detecting multiple errors in a single window. In our variation, we generate a new search space for each window separately. For every word $(w_n)$ in the current window on which the method will run, a set of spelling variations $S_c(w_n)$ is considered. $C(S'')$ will be the search space of the current window, which includes different combinations of the current words and their related spelling variation sets. The sequences in $C(S'')$ are generated with respect to their order of appearance.

Sequences in the current window's search space $C(S'')$ will be the combination of a set of spelling variations and original words, with respect to their order. Figure 1 illustrates a more precise image of the process through equation 3:

(3)
$$\sum_{n=1}^{n=N}\sum_{i=0}^{i=I} W_n C_i$$

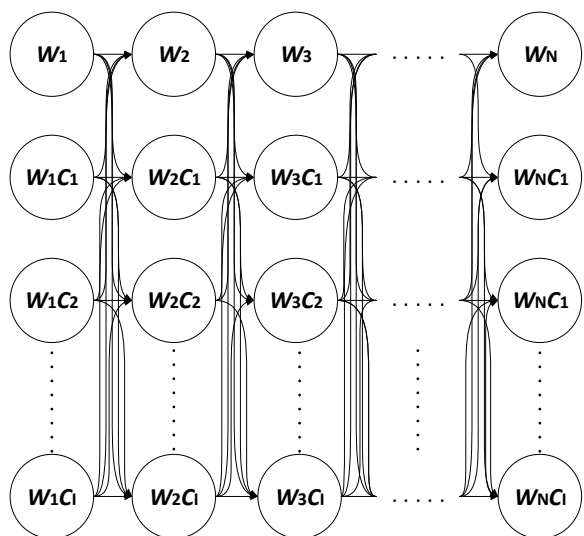

Figure 1: Search space $C(S'')$ for any size of window

According to Figure 1, the first word in every column is the original word in the sentence, with its specific word order typed by the user. Other words in each column shape the related set of spelling variations or $S_c(w_n)$. To be more specific, the search space for the current window, which accommodates $(w_1.w_2.w_3 \ldots .w_n)$ words, includes all of the combinations of the original words and their set of spelling variations with respect to their order of appearance. A combined sequence of words in the current search space is $CS_i$. Figure 1 demonstrates all of the possible combinations in the search space for a fixed-sized window, where $CS_i \in C(S'')$. $W_n C_i$ is the $i$th related spelling variation of the $n$th word in the present window.

### 4.1.2 Using the PCFG

A PCFG encompasses context-free grammars. A probability value is assigned to each production. The probability of a derivation is defined as the product of all the probabilities of the productions used in that specific derivation. These probabilities can be regarded as parameters of the model. The validity of a probabilistic grammar is directly affected by the context of its training dataset. In the task of spelling correction, PCFGs may be used to detect word sequences which are syntactically well formed. Petrov et al. (2006) demonstrated that hierarchically split PCFGs had better accuracy than lexicalized PCFGs. Klein and Manning (2003) observed that lexicalized PCFGs could parse much more accurately than non-lexicalized PCFGs. Additionally, Klein and Manning (2003) developed a high quality lexicalized PCFG which was fed from a tree bank by manual annotation.

In the present study, the Stanford PCFG parser, which was developed based on Klein and Manning's (2003) work, was used to find best potential correction candidates $(CS_i)$ in the current window's search space. The process is accomplished through six steps:

> *Step 1:* The PCFS parser is used to analyze every $CS_i$ in the current search space of the windows. Any $CS_i$ sequence among $C(S'')$, where $CS_i$ shows a parse probability higher than the original words, is stored temporarily as syntactically well-formed candidates.
> *Step2:* Mays and colleagues' (1991) method is run on the words covered by the trigrams that every $CS_i$ among syntactically well-formed candidates contains. In the corpora, those word sequences ($CS_i$) which have equal or higher statistical probabilities in comparison to original sequences of words (user's input) are stored in the windows' candidates list along with their probability values.

*Step 3:* The window moves *d* words to the right and the same process is repeated until all the words in the sentence are covered. In the end, every window in the sentence has its own unique list of probable candidates.

*Step 4:* In the final list, all probable correction candidates with a probability exceeding the value in the original input word sequences are chosen and then combined according to their order of appearance.

*Step 5:* The PCFG is used to calculate the parse probability of each possible combination; combinations with probabilities higher than those of the original sentence will be stored temporarily.

*Step 6:* Through the equation 4, probabilities of the most probable combinations are calculated. $CS_{ij}$ is the *i*th probable word sequence in the *j*th window. In order to find the best combination, the probabilities of the related word sequences (which are members of the windows' candidates list) of a combination are multiplied. The combination with the greatest probability value, compared to that of the original sentence, is chosen as the proper candidate.

(4)
$$P(Combination) = \prod_{i=1}^{n}\prod_{j=1}^{m} P(CS_{ij})$$

What distinguishes this method from those of Mays et al. (1991) and Wilcox-O'Hearn and Hirst (2008) is that the proposed method takes into account the important role of grammatical structure and syntactic knowledge. Unlike the other approaches, the proposed model is not functionally limited to the usage of statistical information. Any potential candidate is verified to be syntactically well-formed. Furthermore, a good candidate recurrently demonstrates good evidence over the corpora or within the list of Ngrams. Candidates with poor grammatical structures are removed from the search space in early stages. As the early steps of applying the PCFG are accomplished, the search space becomes smaller and the speed considerably increases. Using syntactic and statistical knowledge together ensures that the model will have a promising performance in real world applications and daily uses. Figure 2 illustrates the flowchart of the model (see below).

### 4.1.3 A practical example

To explain how the method works in practice, an example is explained here. The following sequence is borrowed directly from 1987–89 Wall Street Journal corpus.

*…the Senate to support aid for the Contras…*

Source of the error: the typist mistakenly typed *aid* as *aim* and *them* as *the*. The task is to detect and correct the misspelled words underlined in the sentence below:

*"BoS...the Senate to support aim for them Contras…EoS"*

First, boundaries of the sentence are specified by using *BoS* and *EoS* markers, so that the window will not move beyond the *EoS* marker. As for the word sequence, assume a window sized 5 (*d*+4), where *d=1*, move on to the words *"to support aim for them."* In the second phase, the search space $C(S'')$ for the current window is generated. Figure 3 shows only part of the initial search space and the possible combinations in it. Then the PCFG is used and parse probability of every $CS_i$ (probable word sequence) is estimated. Any probable word sequence with a higher parse probability is supposed to be syntactically better formed than original word sequences typed by the source of the error. These probable word sequences are temporarily stored. Then, trigrams will be run on the stored word sequence and the product of trigrams will be estimated as the probability of that sequence. The same process will be repeated for every $CS_i$. $CS_i$s with a higher statistical probability in the corpora, compared to original word sequences, are stored in the windows' candidate list.

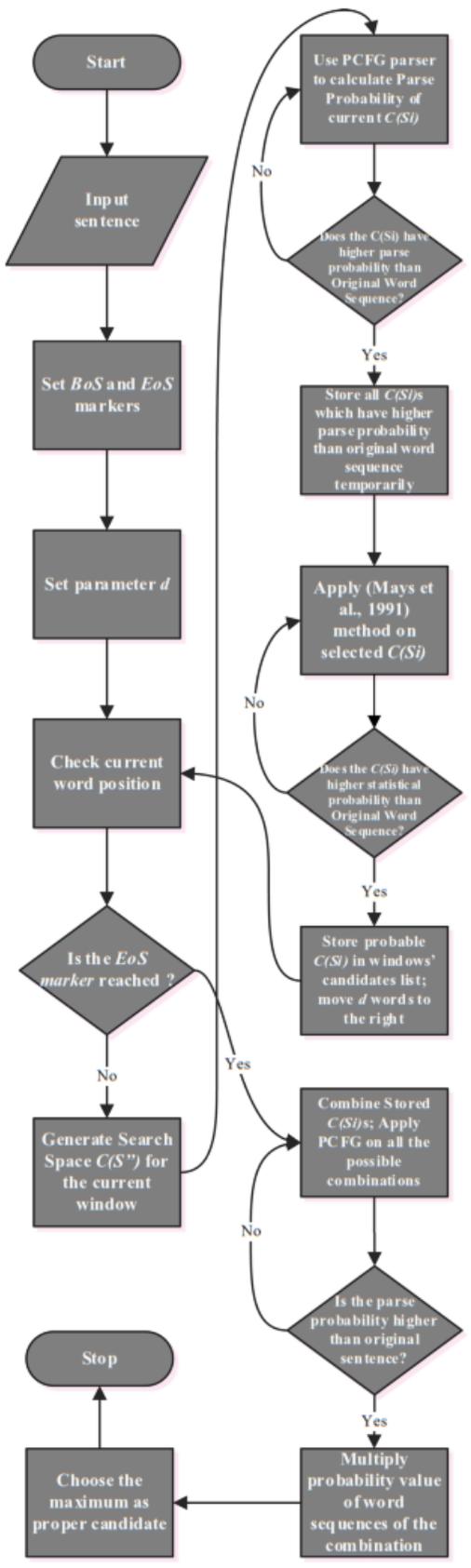

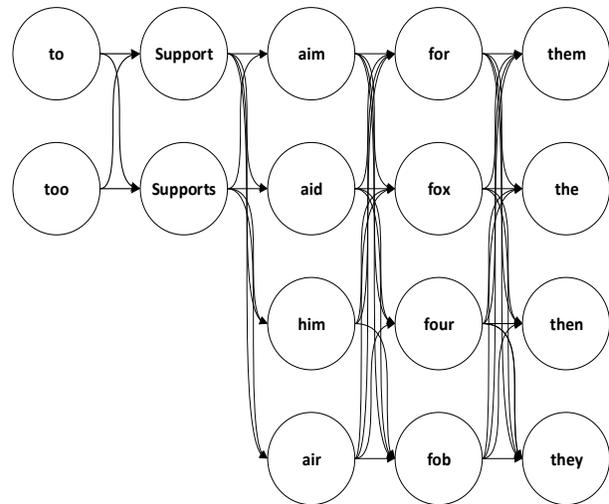

Figure 3: Search space $C(S'')$ for the sequence *"to support aim for them"*

The process reaches closure when the proper word sequences in the first window are selected. The model will continue to move $d$ words to the right in order to cover other words in the sentence. The model moves on to the words *"support aim for them contras."* The search space for this window is represented in Figure 4.

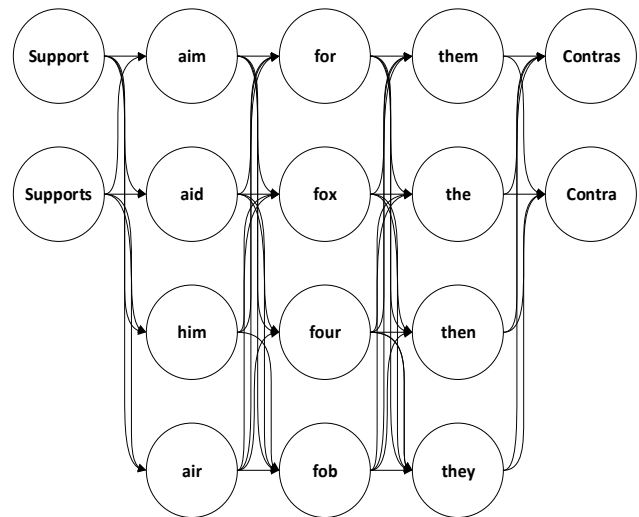

Figure 4: Search space $C(S'')$ for the sequence *"support aim for them Contras"*

All steps are repeated in exactly the same way until the window reaches the *EoS* marker. Then $CS_i$s in the windows' candidates list are combined according to their order of appearance. Next, the PCFG is used to estimate the parse probability of each combination. Combinations with a higher parse probability than the original sentence are stored. Then the statistical probability of each combination is calculated through equation 4. The combination with the highest statistical probability replaces the original sentence. In comparison to Wilcox-O'Hearn and Hirst's (2008) fixed-window size, the proposed model not only allows for a single variation, but it makes it possible to replace multiple variations in the original words in the current window. This would permit multiple corrections in a window. However, in the search space, there are some instances which are grammatically incorrect, and any attempt to estimate the probability of these sequences will slow down the process. Of course, applying this technique to the original sentence in the case of a single large unit will be almost impossible. The reason for this is that the generated combinations would be combinatorially explosive. In order to solve this challenge in this study, the PCFG was manipulated, which proved to be helpful in pruning the search space from syntactically ill-formed candidates.

Figure 2: Flowchart diagram of the model

## 5    Evaluations

In this section, first we discuss some of the main disadvantages of Mays and colleagues' (1991) model. Following that, the approach introduced in this study is evaluated in terms of precision, recall and performance.

### 5.1    Evaluating the model of Mays et al.

The model developed by Mays et al. (1991) was evaluated using trigrams, the vocabulary of which consisted of 20,000 words. Bahl et al. (1983) and Mays et al. (1991) relied on the probabilities derived from IBM Laser Patent Corpus, while this corpus of vocabulary included only 12,000 words. Another drawback of Mays and colleagues' (1991) evaluation was that the test set included 100 sentences which were chosen from Transcripts of the Canadian Parliament and AP newswire (50 from each), although the transcripts are not available today. In Mays and colleagues' (1991) evaluation, for a sentence $S'$ there was a search space of 86 sentences $S$, each containing one error. In their evaluation, no information was provided regarding sentence length. Moreover, the evaluation did not include information about precision and recall. Even the data used is not presently available to calculate the measured values. Considering these drawbacks, the original evaluation presented by Mays et al. (1991) is not reliable, while it cannot be compared with other methods such as those proposed by Hirst and Budanitsky (2005) and Golding and Roth (1999). In the following subsection, a detailed explanation of our evaluation is presented.

### 5.2    Evaluating Multiple Corrections Per Window

Wilcox-O'Hearn and Hirst (2008) proposed a new variation which allowed for multiple corrections and made use of windows of fixed-lengths, while suggesting a new evaluation of Mays and colleagues' (1991) method. Wilcox-O'Hearn and Hirst (2008) followed the work of Hirst and Budanitsky (2005), investigating 1987–89 Wall Street Journal corpus. Following Wilcox-O'Hearn and Hirst (2008), we chose the same corpus, which includes nearly 30 million words in the form of linear text with all headings and identifiers removed. We assumed that the corpus is free from errors. Five hundred articles, each ranging from 90 to approximately 2,700 words (including a total of 300,000 tokens) were put aside to create a test set. Cambridge Statistical Language Modeling Toolkit (Clarkson and Rosenfeld, 1997) was manipulated to create a trigram model. Following Wilcox-O'Hearn and Hirst (2008), a trigram model was created. Wilcox-O'Hearn and Hirst (2008) evaluated the original model using a 20,000 and a 62,000-word corpus. Their evaluations proved that the model, which incorporated 62,000 words, performed significantly better in correcting real- word errors. Following Wilcox-O'Hearn and Hirst (2008), we used as our vocabulary the 62,000 most frequently used words in the remaining texts of the corpus. Using these trigrams, we could directly compare our variation with that of Wilcox-O'Hearn and Hirst's (2008) model. Using the 500 reserved articles, we selected 31110 sentences randomly. Then we created two different test sets, each containing 15,555 sentences described as follows:

> **S62000:** Any word in the corpora of 62,000 frequently used words may be replaced with one of its spelling variations from the same vocabulary.

> **MALP:** Any word whose base form exists as a noun in the knowledgebase of WordNet (regardless of any syntactic analysis) may be replaced by any of its spelling variations.

This would replicate the malapropism data in Hirst and Budanitsky's (2005) model.

|  | Detection | | | Correction | | |
|---|---|---|---|---|---|---|
| α | P | R | F | P | R | F |
| **Hirst and Budanitsky (2005)** | | | | | | |
| Test set **MALP**: | | | | | | |
| .95 | .226 | .311 | .261 | .213 | .287 | .244 |
| **Wilcox-O'Hearn and Hirst, (2008)** | | | | | | |
| Test set **S62000**, $d$=1: | | | | | | |
| .9 | .271 | .862 | .412 | .265 | .834 | .402 |
| .99 | .501 | .777 | .609 | .495 | .762 | .600 |
| .995 | .579 | .753 | .654 | .570 | .736 | .642 |
| .999 | .736 | .678 | .705 | .735 | .675 | .703 |
| Test set **MALP**, $d$=1: | | | | | | |
| .9 | .175 | .610 | .271 | .166 | .581 | .258 |
| .99 | .369 | .545 | .440 | .365 | .527 | .431 |
| .995 | .432 | .513 | .469 | .423 | .496 | .456 |
| .999 | .607 | .447 | .514 | .600 | .439 | .507 |
| **Multiple Corrections per Window** | | | | | | |
| Test set **S62000**, $d$=1: | | | | | | |
| .9 | .404 | .896 | .556 | .388 | .871 | .536 |
| .99 | .517 | .802 | .628 | .516 | .775 | .619 |
| .995 | .584 | .769 | .663 | .576 | .766 | .657 |
| .999 | .738 | .689 | .712 | .737 | .671 | .702 |
| Test set **MALP**, $d$=1: | | | | | | |
| .9 | .294 | .627 | .400 | .285 | .561 | .377 |
| .99 | .385 | .551 | .453 | .378 | .531 | .441 |
| .995 | .446 | .525 | .482 | .447 | .505 | .474 |
| .999 | .611 | .454 | .520 | .611 | .449 | .517 |

Table 1: Comparing results of *multiple corrections per window* with those of Wilcox-O'Hearn and Hirst (2008) and Hirst and Budanitsky (2005) (shown on the first row) using Wall Street Journal corpus including 62,000 words.

We used four different values of α to test the method, from .9, which replicated an unskilled typist, to .999, which simulated a very accurate typist. Depending on α value, different words could be replaced with their spelling variations. For instance, for α =.95, which was the value used by Hirst and Budanitsky (2005), approximately one word in every twenty may be replaced by one of its spelling variations. A spelling variation is defined as any word with the maximum of an edit distance of 1 from the original word, which might be a single insertion, deletion, or substitution, or the transposition of two characters that results in another real-word error. We estimated the results, using three measures: per-word precision, recall, and F1-Score (which is the harmonic mean of recall and precision). F1-Score was calculated through the following formula:

(5)
$$F1 - Score = 2 * \frac{P * R}{P + R}$$

All three measures were demonstrated for the detection and correction of an error. A comparison of results in the case where $d$=1 is demonstrated in Table 1. On both S62000 and MALP test sets, the performance of the model was outstanding. As we expected, with lower values of α, especially when α =.9, precision, recall and F1 score for both detection and correction procedures demonstrated significant increase in comparison to Wilcox-O'Hearn and Hirst's (2008) model. The reason for this is that in contexts where α shows lower values, the probability that a current window might include two or more real-word errors would significantly increase. For higher values of α which are practically more realistic (e.g. α=.995), the proposed method and Wilcox-O'Hearn and Hirst's (2008) model demonstrated almost similar performance. Unsurprisingly, for higher values of α, two real-word errors might rarely appear in a current window. However, results of the MALP test set were noticeably poorer than the S62000 test set; of course, these findings were expectable. Because the MALP test set mainly contained content-word errors, it was less likely for the test to process syntactically ill-formed structures. However, results of the proposed method were obviously better compared to those of Hirst and Budanitsky's (2005) model (see the first row of Table 1). It should be noted that for $d$=1 span of words, where window-

size=5, the search space $C(S'')$ included an average of 10 sequences.

| α | Detection | | | Correction | | |
|---|---|---|---|---|---|---|
| | P | R | F | P | R | F |
| Test set **S62000,** *d*=3: | | | | | | |
| .9 | .411 | .901 | .564 | .399 | .865 | .546 |
| .99 | .523 | .806 | .634 | .520 | .775 | .622 |
| .995 | .587 | .770 | .666 | .581 | .752 | .655 |
| .999 | .742 | .688 | .713 | .737 | .676 | .705 |
| Test set **MALP,** *d*=3: | | | | | | |
| .9 | .301 | .622 | .405 | .295 | .596 | .394 |
| .99 | .386 | .550 | .453 | .383 | .532 | .445 |
| .995 | .453 | .525 | .486 | .443 | .510 | .474 |
| .999 | .616 | .457 | .524 | .611 | .445 | .514 |
| Test set **S62000,** *d*=6: | | | | | | |
| .9 | .421 | .892 | .572 | .408 | .858 | .553 |
| .99 | .529 | .802 | .637 | .527 | .778 | .628 |
| .995 | .592 | .768 | .668 | .578 | .766 | .658 |
| .999 | .750 | .683 | .714 | .743 | .669 | .704 |
| Test set **MALP,** *d*=6: | | | | | | |
| .9 | .306 | .618 | .409 | .300 | .592 | .398 |
| .99 | .393 | .551 | .458 | .388 | .531 | .448 |
| .995 | .459 | .523 | .488 | .441 | .510 | .473 |
| .999 | .618 | .455 | .524 | .613 | .448 | .517 |
| Test set **S62000,** *d*=10: | | | | | | |
| .9 | .433 | .869 | .577 | .415 | .860 | .559 |
| .99 | .545 | .797 | .647 | .536 | .774 | .633 |
| .995 | .599 | .768 | .673 | .589 | .761 | .664 |
| .999 | .755 | .686 | .718 | .747 | .672 | .707 |
| Test set **MALP,** *d*=10: | | | | | | |
| .9 | .319 | .614 | .419 | .307 | .592 | .404 |
| .99 | .399 | .547 | .461 | .393 | .528 | .450 |
| .995 | .461 | .525 | .491 | .457 | .505 | .480 |
| .999 | .621 | .453 | .524 | .616 | .441 | .514 |

Table 2: Using fix-sized windows with higher values of *d*=3 *d*=6 *d*=10, on the S62000 test set.

Afterwards *d* rose up to higher values of 3, 6 and 10. The results, then, were noticeably improved. Table 2 lists the results for d=3, d=6 and d=10. Obviously, with an increase in the values assigned to *d*, the measure of recall decreased. Yet, at the same time, precision noticeably increased. This process yielded significantly higher values for the measure of F1 score. Table 3 represents samples of multiple successful and unsuccessful corrections in a specific window.

| Successful multiple corrections |
|---|
| … *tow* → town [town] saloon after the *battle* → cattle [cattle] roundup. |
| …the Senate to support *aim* → aid [aid] for *them* → the [the] Contras… |
| U.S. manufactures in *sort* → short [short] again are confronting a ball game in *whish* → which [which] they will be able to play. |
| Unsuccessful multiple corrections |
| True Positive Detection of two errors, False Positive Correction of one error: … *worming* → working [working] on improving his *lent* → talent [left]. |

Table 3: Samples of successful and unsuccessful multiple corrections. Italics show the words which are assumed to be errors, arrows show corrections replacing errors, and strings inside brackets demonstrate users' intended words.

The evaluations demonstrated that the method introduced in this study performed well in correcting multiple errors in a specific window. Meanwhile, during the evaluations, as the span of words *d* was expanded and the windows accommodated a greater number of words, the runtime increased as well, and the model underwent some performance overhead in comparison to Wilcox-O'Hearn and Hirst's (2008) model. This situation occurred mainly because the proposed model had to deal with a multitude of combined sequences and to use the PCFG to discriminate between them. The hardware platform which was used to perform this comparison was a HPE ProLiant ML150 Gen9 Server model; with Intel Xeon E5-2600 v4 Processor and 256 GB RAM (DDR4-21400 MHz) installed. Table 4 represents a comparison of average correction times for all the test sentences, using different values of *d* processed through the proposed model and Wilcox-O'Hearn and Hirst's (2008) model.

| $d$ | Correction time (in milliseconds) | |
|---|---|---|
| | Multiple Corrections per Window | Single Correction per Window |
| 1 | 720 | 597 |
| 3 | 925 | 712 |
| 6 | 1488 | 1150 |
| 10 | 2109 | 1473 |

Table 4: Average correction times for all test sentences using different values of *d*

As it was discussed in section 4, for lower values of *d*, more iteration would be required to check a sentence completely. However, the runtime for smaller values of *d* would be significantly better, because in such cases fewer combinations would have to be generated and weeded out.

| $d$ | Initial search space size | Final search space size |
|---|---|---|
| 1 | 154 | 10 |
| 3 | 390 | 19 |
| 6 | 1057 | 22 |
| 10 | 4729 | 39 |

Table 5: Relationship between window size and average size of the search space of a window

Table 5 represents the relationship between window size and average size of the search space of a window. The length of sentences which are used to evaluate the model varies from six to twenty-three words. The average length of sentences is twelve words. The initial search space size represents the number of search space members before the application of the PCFG on average. The final search space size demonstrates the average number of search space members of a window after the PCFG is applied. As the size of the search space grows, the runtime considerably increases. Although the computations are heavy, powerful hardware accomplishes the task noticeably fast. However, in daily applications regular hardware installed on personal computers and mobile devices may have difficulty implementing this model, especially in cases where the size of windows is rather large. Yet, when the model is accessible through service-oriented architecture on powerful servers, web users may have a new experience in detecting and correcting their textual content, with remarkably better accuracy.

## 6   Other Methods

Hirst and Budanitsky (2005) proposed a method for detecting and correcting *malapropism*, which is the use of an incorrect word in a context usually because of phonological confusion. Their method was based on the lexical-resource of WordNet. They used the measure of lexical cohesion to detect semantic distance in the context. If a spelling variation resulted in a word semantically closer to the context, then it was hypothesized that the original word was an error. Nevertheless, their method did not represent good performance in comparison to that of Mays et al. (1991) and Wilcox-O'Hearn and Hirst (2008). Some other recent works, which have focused on correcting real-word errors, have been proposed by Golding and Schabes (1996), Golding and Roth (1996), and Golding and Roth (1999). All of these methods are based on machine learning techniques. They viewed the real-word error correction as a disambiguation task.

Predefined confusion sets are taken mostly from the list of commonly confused words suggested by the Random House Unabridged Dictionary Flexner (1983). This list was used to model anomaly among the words in a context (Golding and Roth, 1996). What distinguished these three methods were the specific techniques they used to address the real-word error correction task. Golding and Roth (1996) and Golding and Roth (1999) used the *WinSpell* software. The method relied on a machine-learning algorithm in which members of confusion sets were demonstrated as clouds of "slow neuron-like" nodes that corresponded to collocation and co-occurrence features. Nevertheless, Golding and Schabes (1996) combined a Bayesian hybrid method, whereas Golding (1995) integrated a *pos*-trigram method. A more recent work has been proposed by Fossati and Di Eugenio (2007, 2008). Their model consisted of a mixed trigram model which used the information of a part of speech tagger. A part of speech tagger determined the grammatical category of each word in a sentence. Mixed trigrams consisted of grammatical categories and/or words, e.g. ("an", adjective, verb). Thus, fewer trigrams were needed. Each word in a sentence was examined to specify the correct grammatical order of the sentence according to the mixed trigrams. The examined word could have several similar candidates (from a confusion set) which had to be checked. In the confusion set, when a word perfectly fitted into the sentence, based on the mixed trigrams, the word would be finally chosen as the correction candidate. Verberne (2002) proposed another method. Her model hypothesized that any word-trigram in the context that had appeared in the BNC (British National Corpus) was correct, and that any trigram that was not present in BNC would definitely be an error. If an unattested word-trigram was observed, then the method tried to use all the possible spelling variations of the current words in the trigram to find attested trigrams. She achieved a recall of .33 and a precision of .05

## 7 Conclusion

We demonstrated that the proposed method outperformed that of Mays et al. (1991) and the variation of trigrams proposed by Wilcox-O'Hearn and Hirst (2008) in terms of correcting multiple errors in a sentence. The model proposed in this study showed better accuracy in correcting multiple errors in a window for lower values of $\alpha$, although for windows with larger sizes, the error correction time increased and performance overhead was noticeable. In the case of correcting malapropisms, following Mays et al. (1991) and Wilcox-O'Hearn and Hirst, (2008), the proposed model performed very well, especially where the number of errors was noticeably higher (lower values of $\alpha$). We succeeded in improving the model by permitting multiple corrections per window, although the runtime was not satisfying. We demonstrated that the introduced variation performed quite well, especially in cases where the source of error (e.g. a typist) was sufficiently skillful.

Yet, correcting multiple errors in a window is a process that can inspire further studies. More specifically, performing precise syntactic analysis on word sequences using English Constraint Grammar (Voutilainen and Heikkilä, 1993) through Mays and colleagues' (1991) model, seems to be an innovative and progressive line of studies. The proposed model may further be improved in terms of performance by using high-speed unlexicalized PCFGs such as one formulated by Petrov et al. (2006). Moreover, one may tradeoff the weight between parse probabilities of syntactic knowledge and statistical probabilities to achieve better results.

## 8 References

Bahl, Lalit R., Frederick Jelinek, and Robert L. Mercer.
  1983. A maximum likelihood approach to continuous speech recognition. IEEE Transactions on Pattern Analysis and Machine Intelligence, 5(2), 179–
  190.

Clarkson, Philip and Roni Rosenfeld. 1997. Statistical language modeling using the CMU–Cambridge Toolkit. Proceedings of the 5th European Conference on Speech Communication and Technology (Eurospeech), Rhodes, 2707–2710.

Flexner, S. B. (editor) 1983. Random House Unabridged Dictionary (2nd ed). Random House.

Fossati, D. and Di Eugenio, B., 2008, May. I saw TREE trees in the park: How to Correct Real-Word Spelling Mistakes. In LREC.

Fossati, D. and Di Eugenio, B., 2007, February. A mixed trigrams approach for context sensitive spell checking. In International Conference on Intelligent Text Processing and Computational Linguistics (pp. 623-633). Springer Berlin Heidelberg.

Golding, A. R. 1995. A Bayesian hybrid method for context-sensitive spelling correction. Proceedings Third Workshop on Very Large Corpora, pp. 39–53. Boston, MA.

Golding, A. R. and Roth, D. 1996. Applying Winnow to context-sensitive spelling correction. In: Saitta, L., editor, Machine Learning: Proceedings 13th International Conference, pp. 182–190. Bari, Italy.

Golding, A. R. and Roth, D. 1999. A Winnow-based approach to context-sensitive spelling correction. Machine Learning, 34(1–3): 107–130.

Golding, A. R. and Schabes, Y. 1996. Combining trigram-based and feature-based methods for context-sensitive spelling correction. Proceedings 34th Annual Meeting of the Association for Computational Linguistics, pp. 71–78. Santa Cruz, CA.

Hirst, G., & Budanitsky, A. 2005. Correcting real-word spelling errors by restoring lexical cohesion. Natural Language Engineering, 11(1), 87-111.

Klein, D. and Manning, C.D., 2003, July. Accurate unlexicalized parsing. In *Proceedings of the 41st Annual Meeting on Association for Computational Linguistics-Volume 1* (pp. 423-430). Association for Computational Linguistics.

Kuenning, G., Willisson, P., Buehring, W. and Stevens, K., 2004. International ispell. *Version*, *3*(00), pp.1-33.

Kukich, K. 1992. Techniques for automatically correcting words in text. ACM Computing Surveys (CSUR), 24(4), 377-439.

Mays, E., Damerau, F. J., & Mercer, R. L. 1991. Context based spelling correction. Information Processing & Management, 27(5), 517-522.

Pedler, J. 2007. Computer correction of real-word spelling errors in dyslexic text. Unpublished PhD thesis. Birkbeck, University of London.

Petrov, S., Barrett, L., Thibaux, R. and Klein, D., 2006, July. Learning accurate, compact, and interpretable tree annotation. In Proceedings of the 21st International Conference on Computational Linguistics and the 44th annual meeting of the Association for Computational Linguistics (pp. 433-440). Association for Computational Linguistics.

Verberne, S. 2002. Context-sensitive spell checking based on word trigram probabilities. Unpublished master's thesis, University of Nijmegen.

Wilcox-O'Hearn, A., Hirst, G., & Budanitsky, A. 2008. Real-word spelling correction with trigrams: A reconsideration of the Mays, Damerau, and Mercer model. In Computational Linguistics and Intelligent Text Processing (pp. 605-616). Springer Berlin Heidelberg.